# An ANN-based Method for Detecting Vocal Fold Pathology

Vahid Majidnezhad
Department of Computer Engineering,
Shabestar Branch, Islamic Azad University,
Shabestar, Iran.

Igor Kheidorov
Department of Computer Engineering,
Shabestar Branch, Islamic Azad University,
Shabestar, Iran.

## ABSTRACT
There are different algorithms for vocal fold pathology diagnosis. These algorithms usually have three stages which are Feature Extraction, Feature Reduction and Classification. While the third stage implies a choice of a variety of machine learning methods, the first and second stages play a critical role in performance and accuracy of the classification system. In this paper we present initial study of feature extraction and feature reduction in the task of vocal fold pathology diagnosis. A new type of feature vector, based on wavelet packet decomposition and Mel-Frequency-Cepstral-Coefficients (MFCCs), is proposed. Also Principal Component Analysis (PCA) is used for feature reduction. An Artificial Neural Network is used as a classifier for evaluating the performance of our proposed method.

## 1. INTRODUCTION
Speech signal information often plays an important role for specialists to understand the process of vocal fold pathology formation. In some cases vocal signal analysis can be the only way to analyze the state of vocal folds. Nowadays diverse medical techniques exist for direct examination and detection of pathologies. Laryngoscopy, electromyography, videokimography are most frequently used by medical specialists. But these methods possess a number of disadvantages. Human vocal tract is hardly-accessible for visual examination during phonation process and that makes it more problematic to identify a pathology. Moreover, these diagnostic means may cause the patients feel much discomfort and distort the actual signal so that it may lead to incorrect diagnosis as well [1-4].

Acoustic analysis as a diagnostic method has no drawbacks, peculiar to the above mentioned methods. It possesses a number of advantages. First of all, acoustic analysis is a non-invasive diagnostic technique that allows pathologists to examine many people in short time period with minimal discomfort. It also allows pathologists to reveal the pathologies on early stages of their origin. This method can be of great interest for medical institutions. In recent years a number of methods were developed for segmentation and classification of speech signals with pathology. The general scheme of vocal fold pathology diagnostic methods consists of three stages which are feature extraction, feature reduction and classification.

Different parameters for feature extraction are used. Traditionally, one deals with such parameters like pitch, jitter, shimmer, amplitude perturbation, pitch perturbation, signal to noise ratio, normalized noise energy [5] and others [6-9]. Feature extraction, using the above mentioned parameters, has shown its efficiency for a number of practical tasks. In the proposed method, we have used the Mel-Frequency-Cepstral-Coefficients (MFCCs), Energy and Shannon Entropy parameters for creating the features vector. Also different approaches for feature reduction are used such as Principal Component Analysis (PCA) [10-13] and Linear Discriminant Analysis (LDA) [14]. In the proposed method we have used PCA for feature reduction. Finally, the reduced features are used for speech classification into the healthy and pathological class. Different machine learning methods such as Support Vector Machines [9], Hidden Markov Model [15], etc can be used as a classifier. In the proposed method we have used the ANN for the classification purpose.

## 2. METHODOLOGY
The wavelet transform, as was shown in [5], is a flexible tool for time-frequency analysis of speech signals. This led us to supposition that feature vectors based on wavelets can show good results. The idea to build feature vector on wavelets for audio classification was previously reported by Li et al [16] and Tzanetakis et al in [17]. These authors used the discrete wavelet transform (DWT) coefficients for their method of feature extraction for content-based audio classification. Kukharchik et al in [18] used continues wavelet transform (CWT) coefficients for their method of feature extraction. Cavalcanti et al in [19] used Wavelet Packet Decomposition (WPD) nodes coefficients for their method for feature extraction. In this paper we have also used the wavelet packet decomposition to create the wavelet packet tree and to extract the features.

The block diagram of our proposed method is illustrated in Fig. 1. In the first stage, by the use of MFCC and Wavelet Packet Decomposition, feature vector containing 139 features is made. In the second stage, by the use of PCA method, the dimension of feature vector from 139 is decreased. In the last stage, by the use of Artificial Neural Network (ANN), the speech signal classified into two classes: pathological or healthy.

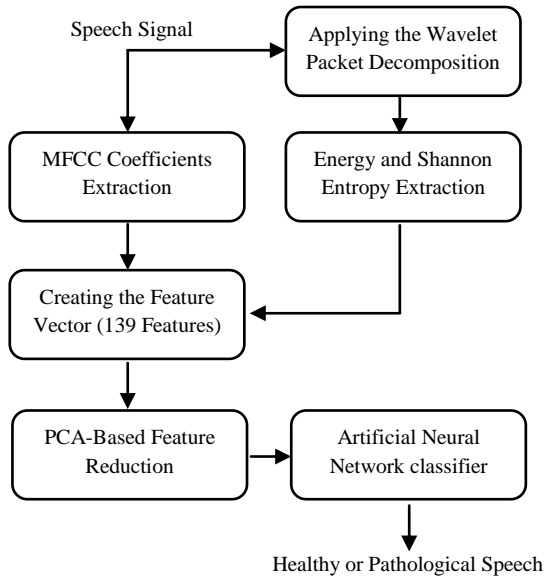

**Fig 1: The scheme of the proposed method for detection of vocal fold pathology.**

## 2.1 Feature extraction
As it is shown in Fig. 1, first, by the use of cepstral representation of input signal, 13 Mel-Frequency-Cepstral-Coefficients (MFCC) are extracted. Then the wavelet packet decomposition in 5 levels is applied on the input signal to make the wavelet packet tree. Then, from the nodes of resulting wavelet packet tree, 63 energy features along with 63 shannon entropy features are extracted. Finally, by the combination of these features, the initial feature vector with the length of 139 features is created.

### 2.1.1 Mel-Frequency-Cepstral-Coefficients
MFCCs are widely used features to characterize a voice signal and can be estimated by using a parametric approach derived from linear prediction coefficients (LPC), or by the non-parametric discrete fast Fourier transform (FFT), which typically encodes more information than the LPC method. The signal is windowed with a hamming window in the time domain and converted into the frequency domain by FFT, which gives the magnitude of the FFT. Then the FFT data is converted into filter bank outputs and the cosine transform is found to reduce dimensionality. The filter bank is constructed using 13 linearly-spaced filters (133.33Hz between center frequencies,) followed by 27 log-spaced filters (separated by a factor of 1.0711703 in frequency.) Each filter is constructed by combining the amplitude of FFT bin. The Matlab code to calculate the MFCC features was adapted from the Auditory Toolbox (Malcolm Slaney). The MFCCs are used as features in [14] to classify the speech into pathology and healthy class. We have used reduction of MFCC information by averaging the sample's value of each coefficient.

### 2.1.2 Wavelet Packet Decomposition
Recently, wavelet packets (WPs) have been widely used by many researchers to analyze voice and speech signals. There are many out-standing properties of wavelet packets which encourage researchers to employ them in widespread fields. The most important, multi resolution property of WPs is helpful in voice signal synthesis [20-21].

The hierarchical WP transform uses a family of wavelet functions and their associated scaling functions to decompose the original signal into subsequent sub-bands. The decomposition process is recursively applied to both the low and high frequency sub-bands to generate the next level of the hierarchy. In this study, mother wavelet function of the tenth order Daubechies has been chosen and the signals have been decomposed to five levels. The mother wavelet used in this study is reported to be effective in voice signal analysis [22-23] and is being widely used in many pathological voice analyses [21]. Due to the noise-like effect of irregularities in the vibration pattern of damaged vocal folds, the distribution manner of such variations within the whole frequency range of pathological speech signals is not clearly known. Therefore, it seems reasonable to use WP rather than DWT or CWT to have more detail sub-bands.

## 2.2 Feature Reduction
Using every feature for classification process is not good idea and it may be causes to the increasing the rate of misclassification. Therefore, it is better to choose the proper features from the whole features. This process is called as "Feature Reduction". One way for feature reduction is Principal Component Analysis (PCA) which is used frequently in pervious works such as [10-13].

PCA method searches a mapping to find the best representation for distribution of data. Therefore, it uses a signal-representation criterion to perform dimension reduction while preserving much of the randomness or variance in the high-dimensional space as possible. The first principal component accounts for as much of the variability in the data as possible, and each succeeding component accounts for as much of the remaining variability as possible. PCA involves the calculation of the eigenvalues decomposition of a data covariance matrix or singular value decomposition of a data matrix, usually after mean centering the data for each attribute. PCA is mathematically defined as an orthogonal linear transformation that transforms the data to a new coordinate system such that the greatest variance by any projection of the data comes to lie on the first coordinate, called the first principal component, the second greatest variance on the second coordinate, and so on.

## 2.3 Artificial Neural Network
An artificial neural network (ANN) as a computing system is made up of a number of simple, and highly interconnected processing elements, which processes information by its dynamic state response to external inputs. In recent times the study of the ANN models is gaining rapid and increasing importance because of their potential to offer solutions to some of the problems which have hitherto been intractable by standard serial computers in the areas of computer science and artificial intelligence. Neural networks are better suited for achieving human-like performance in the fields such as speech processing, image recognition, machine vision, robotic control, etc.

Processing elements in an ANN are also known as neurons. These neurons are interconnected by means of information channels called interconnections. Each neuron can have multiple inputs; while there can be only one output. Inputs to a neuron could be from external stimuli or could be from output of the other neurons. Copies of the single output that comes from a neuron could be input to many other neurons in the network. When the weighted sum of the inputs to the neuron exceeds a certain threshold, the neuron is fired and an output signal is produced. The network can recognize input patterns once the weights are adjusted or tuned via some kind of learning process [24].

## 3. EXPERIMENTS AND RESULTS

The database was created by specialists from the Belarusian Republican Center of Speech, Voice and Hearing Pathologies. We have selected 75 pathological speeches and 55 healthy speeches randomly which are related to sustained vowel "a". All the records are wave files in PCM format. The whole scheme of our proposed method is illustrated in Fig. 1. We have adopted a 10 fold cross-validation scheme to assess the generalization capabilities of the system in our experiments.

It is necessary to know how many neurons in the hidden layer are required to achieve the optimal results. So, in the first experiment the numbers of neurons in the hidden layer are investigated and the results are shown in Fig. 2. As it is obvious, using 5 neurons in the hidden layer lead to the optimal case which is 84.62% of accuracy.

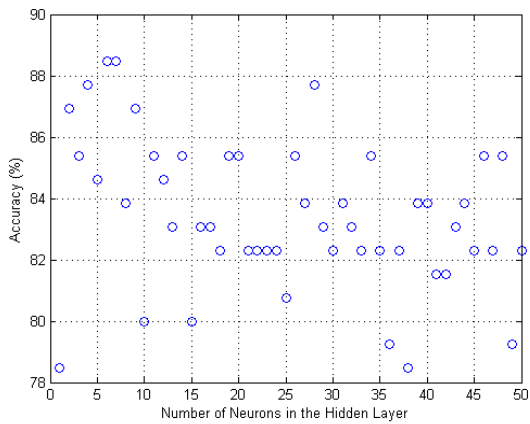

**Fig 2: The Classification Results based on the numbers of Neurons.**

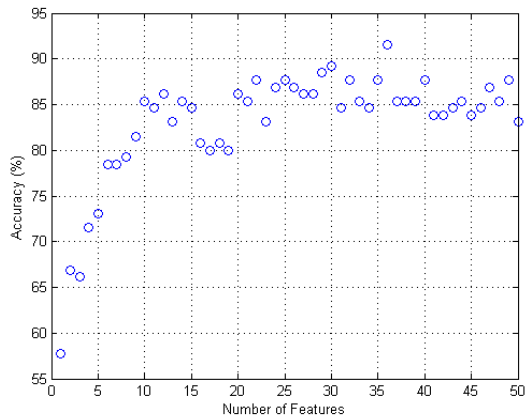

**Fig 3: The Classification Results based on the numbers of Features.**

Finally, according to the result of the first experiment, an ANN with 5 neurons is created for using in the second experiment. The goal of the second experiment is to reduce the initial feature vector's length so that the accuracy of the classification phase increases. For this purpose, the conventional PCA method is used and the results are shown in Fig. 3. As it is obvious, using the 36 features lead to the best result which is 91.54% of accuracy. The 36 selected features, by the means of PCA method, are shown in the table 1.

**Table 1. The selected features for the construction of feature vector.**

| Feature Reduction Method | The selected features | Accuracy |
|---|---|---|
| PCA (feature vector length=36) | The 1-8 and 10-13 coefficients of MFCC. Energy at the 9-10, 17, 19, 22-23, 32, 35-38 and 46th nodes of WP Tree. Entropy at the 1, 4, 9, 16-17, 19, 22, 33-34, 37, 39 and 45th nodes of WP Tree. | 91.54% |
| None (feature vector length=139) | All the 63 entropy & 63 energy & 13 MFCC coefficients. | 84.62% |

## 4. CONCLUSION AND FUTURE WORKS

Acoustic analysis is a proper method in vocal fold pathology diagnosis so that it can complement and in some cases replace the other invasive, based on direct vocal fold observation, methods. In this article, an ANN-Based method for vocal fold pathology diagnosis is proposed so that in the proposed scheme, Mel-Frequency-Cepstral-Coefficients along with the wavelet packet decomposition are used for feature extraction phase. Also PCA method for the feature reduction phase is used. And finally the ANN is used for the classification phase. Two experiments are designed to investigate the optimal case for the numbers of neurons in the hidden layer of the ANN and also the optimal case for the feature vector length as the input of ANN.

Also it may be possible to try to build a complete multiclass classification system so that detection of different type of pathological speech will be possible. For this propose, we suppose that further research for more sophisticated feature extraction phase.

## 5. ACKNOWLEDGMENTS

This work was supported by the speech laboratory of the United Institute of Informatics Problems of NASB. The authors wish to thank the Belarusian Republican Center of Speech, Voice and Hearing Pathologies by its support in the speech database.